\documentclass[runningheads]{llncs}
\usepackage{graphicx}
\usepackage{siunitx}
\usepackage{amssymb}

\begin{document}
\title{Deep Semi-Supervised and Self-Supervised Learning for Diabetic Retinopathy Detection}

\author{Jose Arrieta\inst{1} \and
Oscar J. Perdomo\inst{2}\and
Fabio A. Gonz\'{a}lez\inst{1}}

\authorrunning{J. Arrieta et al.}% 

\titlerunning{Deep Semi-Supervised and Self-Supervised Learning for DR Detection}%


\institute{MindLab Research Group, Universidad Nacional de Colombia,  Colombia
\email{\{jmarrietar\inst{1}, ojperdomoc\inst{2}, fagonzalezo\inst{3}\}@unal.edu.co}
\and
Universidad del Rosario, Colombia\\}

\maketitle              % typeset the header of the contribution
\begin{abstract}
Diabetic retinopathy (DR) is one of the leading causes of blindness in the working-age population of developed countries, caused by a side effect of diabetes that reduces the blood supply to the retina. Deep neural networks have been widely used in automated systems for DR classification on eye fundus images. However, these models need a large number of annotated images. In the medical domain, annotations from experts are costly, tedious, and time-consuming; as a result, a limited number of annotated images are available. This paper presents a semi-supervised method that leverages unlabeled images and labeled ones to train a model that detects diabetic retinopathy. The proposed method uses unsupervised pretraining via self-supervised learning followed by supervised fine-tuning with a small set of labeled images and knowledge distillation to increase the performance in classification task. This method was evaluated on the EyePACS test and Messidor-2 dataset achieving 0.94 and 0.89 AUC respectively using only 2\% of EyePACS train labeled images.

\keywords{Diabetic Retinopathy  \and Medical Imaging \and Deep Learning \and Semi-supervised Learning \and Self-supervised learning.}

\end{abstract}
\section{Introduction}
\label{sec:intro}

Diabetic retinopathy (DR) is one of the leading causes of blindness in the working-age population in developed countries~\cite{Organisation2019}. DR consists of a side effect of diabetes that reduces the blood supply to the retina, including lesions that appear on the surface, such as microaneurysms, exudates, hemorrhages, and cotton wool spots. Automated methods for detecting eye diseases present as an useful tool for early diagnosis, important to prevent the occurrence of blindness and lack of vision \cite{Bhaskaranand2019,Beede2020}. Recently, deep learning methods became popular for DR classification on eye fundus images because of their promising results~\cite{Asiri2019,Gulshan2016,Voets2019,Krause2018,Quellec2017,Rakhlin2018,Toledo-cort}. However, the majority of these approaches use labeled images, whereas the manual labeling of medical images is expensive and time-consuming because it requires medical experts in the retina. As a result, relatively small labeled data sets are available to train deep learning models.

\begin{figure}[htb]
\begin{minipage}[b]{1.0\linewidth}
  \centering
  \centerline{\includegraphics[width=12cm]{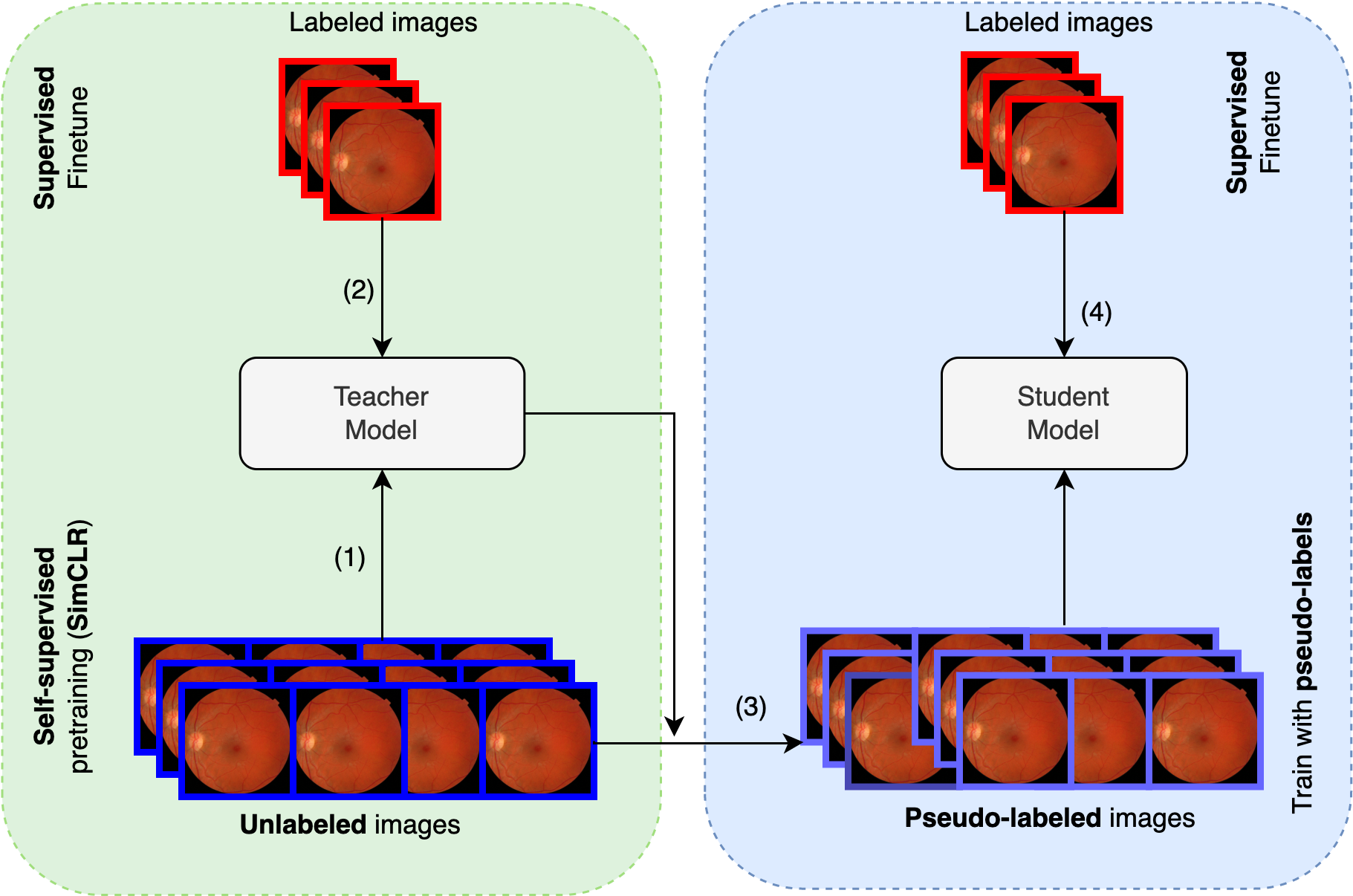}}
%  \vspace{2.0cm}
\end{minipage}
\caption{The proposed method consists of four stages as follows: (1) Self-supervised pre-training using SimCLR with unlabeled images. (2) Supervised fine-tuning on a small set of labeled images. (3) Knowledge distillation from teacher model to student model using teacher's output as pseudo-labels to train student model. (4) Fine-tuning of student model on a small set of labeled images.}
\label{fig:Pipeline}
\end{figure}

Some authors have proposed semi-supervised learning approaches to leverage unlabeled images and mitigate the lack of annotated images. In specific, Liu et al.~\cite{Liu2019} and Xie et al.~\cite{Xie} propose the use of generative adversarial networks (GAN) to extend classification while performing unsupervised image reconstruction with unlabeled data in training. State-of-the-art methods in deep semi-supervised methods such as MixMatch~\cite{Berthelot2019} and, FixMatch~\cite{Sohn} comprise strong and weak augmentations of unlabeled images and make class pseudo-labels match via consistency regularization. In medical imaging, Hansen et al.~\cite{Hansen2020} explore MixMatch in two distinct medical imaging domains as skin lesion diagnosis and lung cancer prediction. Pooch et al.~\cite{Pooch2020} evaluate the performance of different methods including pseudo-labeling, Mean Teacher~\cite{Tarvainen2017}, Unsupervised Data Augmentation (UDA)~\cite{Xie2020}, MixMatch~\cite{Berthelot2019} and FixMatch~\cite{Sohn} in a chest radiography classification task. 

Alternatively, self-supervised learning \cite{Chen2019} presents as a new strategy to use unlabeled data to pre-train a neural network and allows the construction of models that can learn relevant image representations from unlabeled medical images. These  visual representations from unlabeled data are used to train strong semi-supervised models~\cite{Learners2020}.  Azizi et al.~\cite{Azizi} presented the effectiveness of self-supervised learning with unlabeled data for domain-specific medical images in dermatology skin and chest X-ray classification. Vu et al. \cite{Vu2021} use self-supervised contrastive learning and propose to use patient metadata to select the pairs of medical images requiring them to be from the same patient. Kaku et al. \cite{Kaku2021}, minimize the mean squared error between the intermediate layer representations in complement to the contrastive loss using medical datasets such as NIH-Chest X-rays, breast cancer histopathology, and diabetic retinopathy.

This paper presents a method that combines self-supervised learning along with a teacher-student and knowledge distillation strategies to perform semi-supervised learning with a small subset of labeled eye fundus images. The method achieves 0.94 AUC on EyePACS test and 0.89 AUC on Messidor-2 Diabetic Retinopathy dataset using only 2\% EyePACS-Kaggle train labeled images (\num{1000} images).

\section{Methods}
\label{sec:Methods}

The proposed method consists of a teacher and a student network. Our method differs from previous approaches because the teacher network heavily relies on large unlabeled diabetic retinopathy (DR) images to learn useful representations in pre-training compared to the labeled ones available. Furthermore, the student network benefits from pseudo-labels created with the teacher model using unlabeled images. The complete semi-supervised workflow proposed is summarized in Figure \ref{fig:Pipeline} and consists of four stages.\\

\begin{figure}[t]
  \centering
   %\includegraphics[width=0.8\linewidth]{egfigure.eps}
   \centerline{\includegraphics[width=1.0\linewidth]{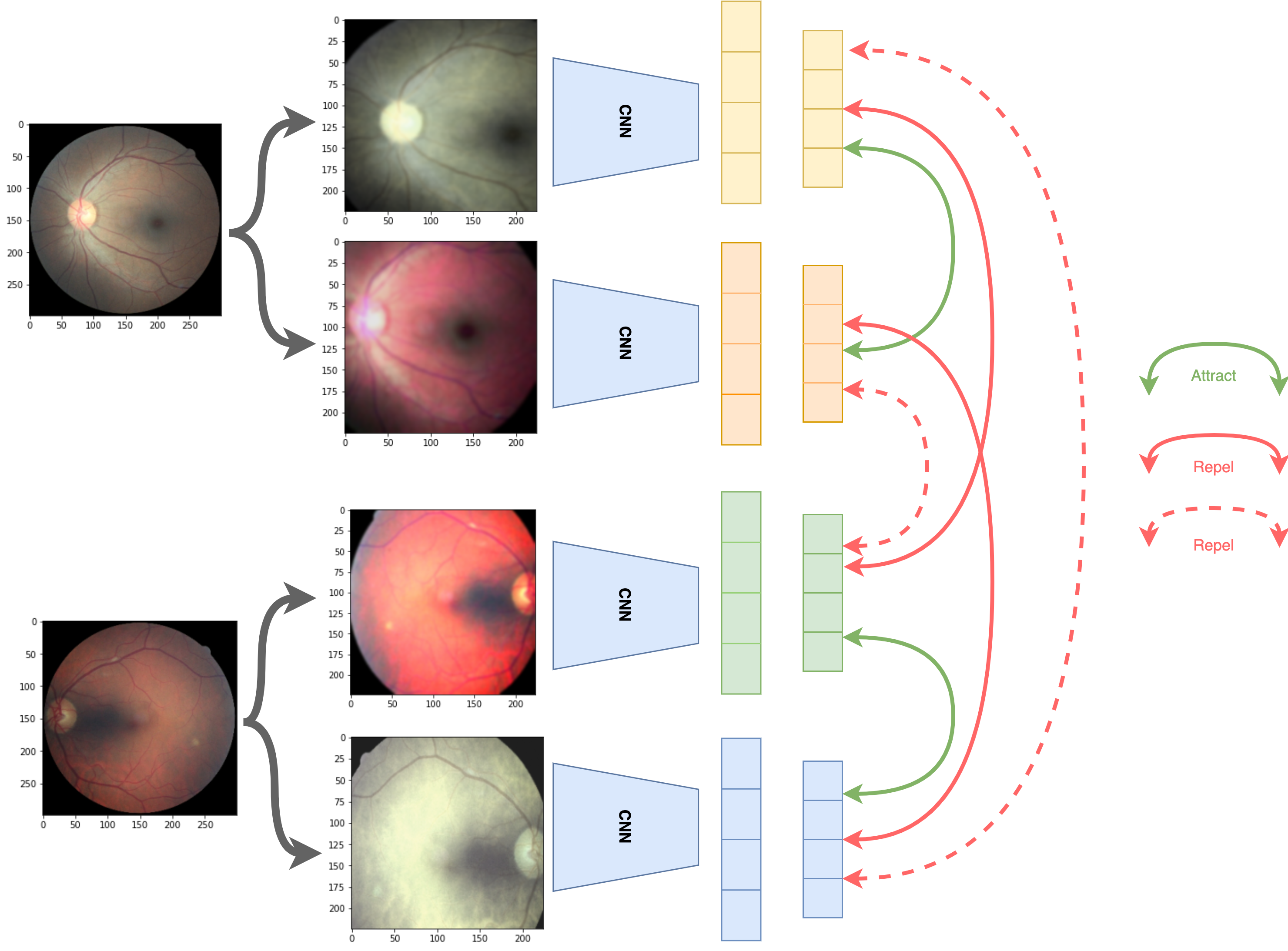}}
    \caption{Self-supervised pre-training with SimCLR uses strong data augmentations to maximize the agreement between different representations of the same image in the latent space.} \label{fig:SimCLR}
\end{figure}

\textbf{1) Self-supervised unlabeled pre-training:} First; the Teacher model is pretrained using the simple framework for contrastive learning of visual representation (SimCLR \cite{Chen2019}). This process is unsupervised as it doesn't need the labels of the images. The whole EyePACS-Kaggle train dataset (57 146 images) without the labels and a ResNet-50 backbone were used to learn useful visual representations with domain-specific medical images. SimCLR uses strong data augmentations such as color distortions (brightness, contrast, saturation, hue), cropping and rotations as shown in Figure \ref{fig:SimCLR}. Contrastive loss \textit{NT-Xent} (the normalized temperature-scaled cross-entropy loss) is used to maximize the agreement of diverse representations of the same image in a latent space, defined as:
\begin{equation}
\mathbb{L}_{i,j} = -\log\frac{\exp\left(\text{sim}\left(\mathbf{z}_{i}, \mathbf{z}_{j}\right)/\tau\right)}{\sum^{2N}_{k=1}\exp\left(\text{sim}\left(\mathbf{z}_{i}, \mathbf{z}_{k}\right)/\tau\right)}
\end{equation}
where $\mathbf{z}_{i}$ and $\mathbf{z}_{j}$ correspond to
the outputs of the same image from the non-linear projection head and $\mathbf{z}_{k}$ corresponds to the remainder of representations of the other $\mathbf{N}$ images in the batch. Additionally, \begin{math}\tau\end{math} is the temperature parameter t, and \textit{sim} denotes a distance metric.\\

\textbf{2) Fine-tuning of Teacher model with labeled images}: In this phase, a small set of 1000 random samples of labeled fundus images are used to further refine the weights of the model. Fine-tuning is a common approach where a network from a previous task is used as starting point, and weights are adjusted to fit a new specific task. The proposed approach used the previously pretrained SimCLR ResNet-50 \cite{Xie2017} model with an additional sigmoid layer at the top to make a binary classification from referable DR and non-referable DR.\\

\textbf{3) Pseudo-labeling and knowledge distillation:} The third step consists of knowledge distillation from the ResNet-50 model (Teacher) to a DenseNet161 model (Student) using unlabeled data as input and teacher's output as pseudo-labels. Knowledge distillation is the process of transferring knowledge from one model to another one, originally used to compress a large model to a smaller one, encouraging the student to match a teacher's output~\cite{Hinton2015}. However, we chose a DenseNet161, a deeper network than the teacher for Student architecture because Xie et al.~\cite{Xie2020}  demonstrated that using student models that are equal to or larger than the teacher in addition to some noise to the student can improve the performance of the student beyond the teacher, making predictions with more difficult images by giving the student model enough capacity and difficult images to learn through. We transform soft-pseudo-labels to hard-pseudo-labels using a threshold of 0.5.\\

\textbf{4) Fine-tuning of Student model with labeled images:} Finally, the fourth task consists of a final fine-tuning of the student model with the same set of 1000 labeled eye fundus images.

\section{Experimental Evaluation}
\label{sec:experimental_evaluation}

\subsection{Datasets}

\begin{figure}[htb]
\begin{minipage}[b]{.48\linewidth}
  \centering
  \centerline{\includegraphics[width=4.0cm]{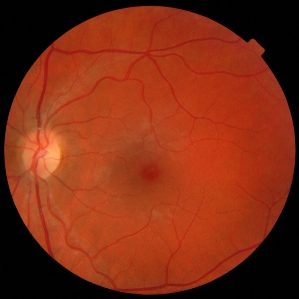}}
%  \vspace{1.5cm}
\end{minipage}
\hfill
\begin{minipage}[b]{0.48\linewidth}
  \centering
  \centerline{\includegraphics[width=4.0cm]{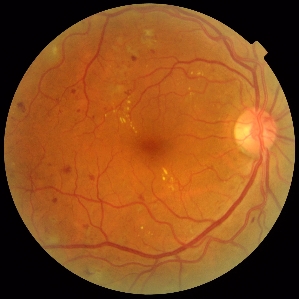}}
%  \vspace{1.5cm}
\end{minipage}
\caption{Example of eye fundus image (EFI) non-referable DR on the left and a referable DR example on the right.}
\label{fig:res}
\end{figure}

EyePACS-Kaggle \cite{kaggle_2015} consists of retina images taken under a variety of imaging conditions as shown in figure \ref{fig:res}, where the clinician rate the presence of diabetic retinopathy in each image on a scale of 0 to 4 (No DR, Mild, Moderate, Severe, Proliferative), respectively. For the binary classification task, 
grades 0 and 1 correspond to non-referable DR, while grades 2, 3, and 4 correspond to referable DR according to the International Clinical Diabetic Retinopathy Scale \cite{Scale2002}. The EyePACS-Kaggle dataset is partitioned into train and tests datasets, with 35121 images and \num{42918} images respectively of different sizes. In this work, we used the EyePACS-Kaggle partition defined in Voets et al. \cite{Voets2019} which consists of a train set of \num{57146} images and a test set of \num{8790} images. Training is performed on the EyePACS-Kaggle train and evaluation is performed on the EyePACS-Kaggle test and Messidor-2 datasets. Messidor-2 \cite{Messidor} consists of a public dataset that contains \num{1748} images with grades adjudicated by a panel of three retina specialists and constitutes a standard dataset used to compare performance results in Diabetic Retinopathy detection. Following Voets et al. \cite{Voets2019} and Gulshan et al. \cite{Gulshan2016} pre-processing, all fundus images are centered and resized to 299 x 299 pixels, with the fundus center in the middle of the image.

\subsection{Experimental setup}
All models were implemented using Pytorch \cite{NEURIPS2019_9015}. EyePACS-Kaggle train dataset without labels is used for self-supervised pretraining a ResNet-50 network using SimCLR data augmentations alongside a projection head with a contrastive loss function, a learning rate of \num[round-precision=2,round-mode=figures, scientific-notation=true]{0.00001}, weight decay of \num[round-precision=2,round-mode=figures, scientific-notation=true]{0.0005} and batch of size 64 for 100 epochs. For knowledge distillation, the previously trained ResNet-50 network is used as a teacher to predict pseudo-labels of the EyePACS-Kaggle train dataset while using them to train a different architecture. DenseNet161, a deeper network is used to improve the performance of the student beyond the teacher. The data augmentation configuration consists of random color jitter and random horizontal flips with BinaryCrossEntropy loss function and stochastic gradient descent as an optimizer, a learning rate of \num[round-precision=2,round-mode=figures,scientific-notation=true]{0.0001} and batch size 32 for 200 epochs. A final fine-tuning of the student network is performed using the 1000 images labeled data for 100 epochs.

Three baselines were defined. First, the supervised Inception-V3 network as proposed in Voets et al. \cite{Voets2019} and Krause et al. \cite{Krause2018}, although some other architectures were also tested with similar performance. Additionally, two state-of-the-art hybrid methods, such as MixMatch and FixMacth, were chosen to compare the performance of semi-supervised approaches.

%------------------------------------------------------------------------
\section{Results}
\label{sec:results}

\begin{figure}[t]
  \centering
   %\includegraphics[width=0.8\linewidth]{egfigure.eps}
   \centerline{\includegraphics[width=1.05\linewidth]{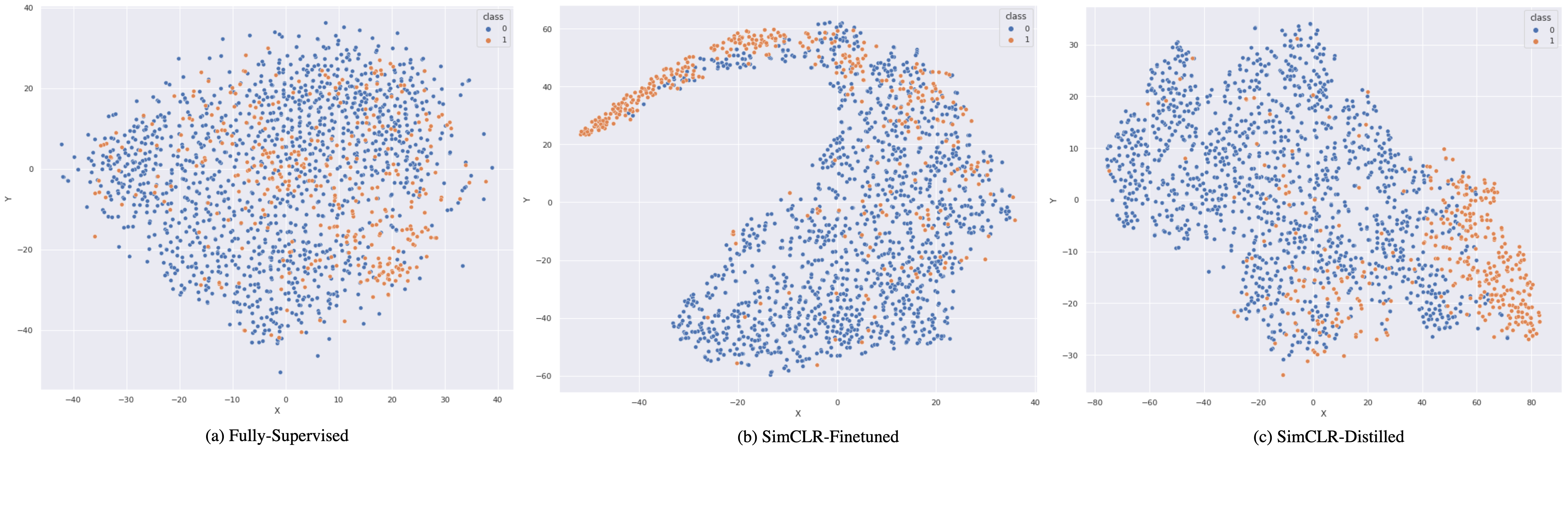}}
    \caption{t-SNE visual representation in Messidor-2: (a) Supervised, (b) SimCLR-Finetuned (Teacher), and (c) SimCLR-Distilled methods (Student).} \label{fig:t-SNE}
\end{figure}

The proposed method is compared to a supervised baseline that uses the fraction of 1000 labeled images (2\% of EyePACS-Kaggle) and semi-supervised learning methods such as FixMatch and MixMatch that make class pseudo-labels match via consistency
regularization using additional unlabeled data. Additionally, a supervised method with the complete train dataset is used as a reference for performance. The proposed method successfully
outperforms the methods trained with a labeled fraction of 2\% of EyePACS train and its at similar performance compared to 100\% of train labeled images method leveraging the use of unlabeled data in the pre-train stage and creating useful visual representations. Additionally, the knowledge distillation step creates a boosting under the AUC metric as reported in Table \ref{tab:tab1} completing a better generalization task of detection of diabetic retinopathy using only 1000 labeled images 2\% of EyePACS-Kaggle and the rest as unlabeled images. Further, in Figure \ref{fig:t-SNE} the representation of visual features is presented using the t-SNE technique in the Messidor-2 dataset. The supervised representation of DR and non-DR seems not easily separable. In contrast, the SimCLR-Finetuned (Teacher) and SimCLR-Distilled (Student) seems more separable between the DR and non-DR instances, improving visual representations and generalization.

\begin{table}
  \centering
\caption{AUC performance measures of baselines and proposed method on test datasets. Supervised baseline is trained using 2\% of EyePACS train labeled images. Semi-supervised methods use 2\% of EyePACS train labeled images and the rest as unlabeled images.}\label{tab:tab1}
\begin{tabular}{|c|l|c|c|c|}
\hline

Label fraction & Method & Architecture & Kaggle Test & Messidor2\\
\hline
100\% & Supervised  &  InceptionV3  &  0.96 & 0.88 \\
\hline
    & Supervised & InceptionV3 &  0.88 & 0.66\\
    & MixMatch \cite{Berthelot2019}  & ResNet50 & 0.85  & 0.64\\
2\% & FixMatch \cite{Sohn}  & ResNet50  & 0.83  & 0.79 \\
\textit{Train} & SimCLR-Finetuned (Teacher) & ResNet50 & 0.92 & 0.85 \\
    & SimCLR-Distilled (Student) & DenseNet161  & \textbf{0.94} &  \textbf{0.89} \\
\hline
\end{tabular}
\end{table}

%-------------------------------------------------------------------------

\section{Discussion and Conclusion}
\label{sec:discussion}

Despite the deep learning model's need for a large amount of labeled data for their training process, this paper presents a method that combines self-supervised learning along with a teacher-student and knowledge distillation strategies to perform semi-supervised learning with a small subset of labeled eye fundus images. The method achieves 0.94 AUC on the EyePACS test and 0.89 AUC on Messidor-2 Diabetic Retinopathy dataset using only 2\% EyePACS train labeled images improving the performance from the baseline supervised approach, and with greater performance than the other two state-of-the-art methods, alongside a better generalization compared to the teacher model. Self-supervised learning resulted in a great pre-training strategy to use unlabeled medical images and create richer and better visual representations improving downstream tasks such as diabetic retinopathy detection in this case. Although one disadvantage of this approach consists of computing and memory intensive, as it creates several representations from each image in the training batch and could require a lengthy training process. Also, knowledge distillation from a teacher to a student model improved performance in the presented results. Furthermore, the student network benefits from pseudo-labels created with the teacher model using unlabeled images, improving the generalization of outcomes and enhancing the performance beyond the teacher. In future works, the validation of our proposed model in a clinical environment, combining images from different sources, equipment, and audiences, especially in the unsupervised steps. Additionally, it was hypothesized that this method could be applied to other medical image domains where labeled images are scarce and constitute a vital research topic for deep learning models. 

\bibliographystyle{plain}
\bibliography{egbib}

\begin{thebibliography}{10}

\bibitem{Asiri2019}
Norah Asiri, Muhammad Hussain, Fadwa Al, and Nazih Alzaidi.
\newblock {Deep learning based computer-aided diagnosis systems for diabetic
  retinopathy : A survey}.
\newblock {\em Artificial Intelligence In Medicine}, 99(December 2018):101701,
  2019.

\bibitem{Azizi}
Shekoofeh Azizi, Basil Mustafa, Fiona Ryan, Zachary Beaver, Jan Freyberg,
  Jonathan Deaton, Aaron Loh, Alan Karthikesalingam, Simon Kornblith, Ting
  Chen, Vivek Natarajan, and Mohammad Norouzi.
\newblock {Big Self-Supervised Models Advance Medical Image Classification}.
\newblock pages 1--19.

\bibitem{Beede2020}
Emma Beede, Elizabeth Baylor, Fred Hersch, Anna Iurchenko, Lauren Wilcox,
  Dr.~Paisan Raumviboonsuk, and Laura Vardoulakis.
\newblock {A Human-Centered Evaluation of a Deep Learning System Deployed in
  Clinics for the Detection of Diabetic Retinopathy}.
\newblock {\em CHI 2020 Paper}, pages 1--12, 2020.

\bibitem{Berthelot2019}
David Berthelot, Avital Oliver, Nicholas Carlini, Ian Goodfellow, Colin Raffel,
  and Nicolas Papernot.
\newblock {MixMatch : A Holistic Approach to Semi-Supervised Learning}.
\newblock (NeurIPS):1--11, 2019.

\bibitem{Bhaskaranand2019}
Malavika Bhaskaranand, Chaithanya Ramachandra, Sandeep Bhat, Jorge Cuadros,
  Muneeswar~G. Nittala, Srinivas~R. Sadda, and Kaushal Solanki.
\newblock {The value of automated diabetic retinopathy screening with the
  EyeArt system: A study of more than 100,000 consecutive encounters from
  people with diabetes}.
\newblock {\em Diabetes Technology and Therapeutics}, 21(11):635--643, 2019.

\bibitem{Chen2019}
Ting Chen, Simon Kornblith, Mohammad Norouzi, and Geoffrey Hinton.
\newblock {A Simple Framework for Contrastive Learning of Visual
  Representations}.
\newblock 2019.

\bibitem{Learners2020}
Ting Chen, Simon Kornblith, Kevin Swersky, Mohammad Norouzi, and Geoffrey
  Hinton.
\newblock {Big Self-Supervised Models are Strong Semi-Supervised Learners}.
\newblock (NeurIPS):1--18, 2020.

\bibitem{Messidor}
Laboratoire de~Traitement de~l'Information Médicale.
\newblock Messidor-2 dataset.
\newblock https://www.adcis.net/en/third-party/messidor2/, 2011.

\bibitem{Gulshan2016}
Varun Gulshan, Lily Peng, Marc Coram, Martin~C Stumpe, Derek Wu, Arunachalam
  Narayanaswamy, Subhashini Venugopalan, Kasumi Widner, Tom Madams, Jorge
  Cuadros, Ramasamy Kim, Rajiv Raman, Philip~C Nelson, Jessica~L Mega, and
  Dale~R Webster.
\newblock {Development and Validation of a Deep Learning Algorithm for
  Detection of Diabetic Retinopathy in Retinal Fundus Photographs}.
\newblock 94043:1--9, 2016.

\bibitem{Hansen2020}
Colin~B. Hansen, Vishwesh Nath, Riqiang Gao, Camilo Bermudez, Yuankai Huo,
  Kim~L. Sandler, Pierre~P. Massion, Jeffrey~D. Blume, Thomas~A. Lasko, and
  Bennett~A. Landman.
\newblock {Semi-supervised Machine Learning with MixMatch and Equivalence
  Classes}.
\newblock {\em Lecture Notes in Computer Science (including subseries Lecture
  Notes in Artificial Intelligence and Lecture Notes in Bioinformatics)}, 12446
  LNCS:112--121, 2020.

\bibitem{Hinton2015}
Geoffrey Hinton, Oriol Vinyals, and Jeff Dean.
\newblock {Distilling the Knowledge in a Neural Network}.
\newblock pages 1--9, 2015.

\bibitem{Kaku2021}
Aakash Kaku, Sahana Upadhya, and Narges Razavian.
\newblock Intermediate layers matter in momentum contrastive self supervised
  learning.
\newblock 2021.

\bibitem{Krause2018}
Jonathan Krause, Varun Gulshan, Ehsan Rahimy, Peter Karth, Kasumi Widner,
  Greg~S. Corrado, Lily Peng, and Dale~R. Webster.
\newblock {Grader Variability and the Importance of Reference Standards for
  Evaluating Machine Learning Models for Diabetic Retinopathy}.
\newblock {\em Ophthalmology}, 125(8):1264--1272, 2018.

\bibitem{Liu2019}
Sijie Liu, Jingmin~Xin B, Jiayi Wu, and Peiwen Shi.
\newblock {Semi-supervised Adversarial Learning for Diabetic Retinopathy
  Screening}.
\newblock pages 60--68, 2019.

\bibitem{Organisation2019}
World~Health Organisation.
\newblock {\em {World report on vision. 2019}}, volume 214.
\newblock 2019.

\bibitem{NEURIPS2019_9015}
Adam Paszke, Sam Gross, Francisco Massa, Adam Lerer, James Bradbury, Gregory
  Chanan, Trevor Killeen, Zeming Lin, Natalia Gimelshein, Luca Antiga, Alban
  Desmaison, Andreas Kopf, Edward Yang, Zachary DeVito, Martin Raison, Alykhan
  Tejani, Sasank Chilamkurthy, Benoit Steiner, Lu~Fang, Junjie Bai, and Soumith
  Chintala.
\newblock Pytorch: An imperative style, high-performance deep learning library.
\newblock In {\em Advances in Neural Information Processing Systems 32}, pages
  8024--8035. Curran Associates, Inc., 2019.

\bibitem{Pooch2020}
Eduardo~H.P. Pooch, Pedro Ballester, and Rodrigo~C. Barros.
\newblock {Semi-supervised Classification of Chest Radiographs}.
\newblock {\em Lecture Notes in Computer Science (including subseries Lecture
  Notes in Artificial Intelligence and Lecture Notes in Bioinformatics)}, 12446
  LNCS:172--179, 2020.

\bibitem{Quellec2017}
Gwenol{\'{e}} Quellec, Katia Charri{\`{e}}re, Yassine Boudi, and B{\'{e}}atrice
  Cochener.
\newblock {Deep image mining for diabetic retinopathy screening}.
\newblock 39:178--193, 2017.

\bibitem{Rakhlin2018}
Alexander Rakhlin.
\newblock {Diabetic Retinopathy detection through integration of Deep Learning
  classification framework}.
\newblock pages 1--11, 2018.

\bibitem{kaggle_2015}
Diabetic retinopathy detection.
\newblock Eyepacs.
\newblock www.kaggle.com/competitions/diabetic-retinopathy-detection, 2015.

\bibitem{Scale2002}
International Clinical Diabetic Retinopathy Disease~Severity Scale.
\newblock {International Clinical Diabetic Retinopathy}.
\newblock (October):8500, 2002.

\bibitem{Sohn}
Kihyuk Sohn, David Berthelot, Chun-liang~Li Zizhao, Zhang Nicholas, Ekin~D
  Cubuk, Alex Kurakin, Han Zhang, and Colin Raffel.
\newblock {FixMatch: Simplifying Semi-Supervised Learning with Consistency and
  Confidenc}.

\bibitem{Tarvainen2017}
Antti Tarvainen and Harri Valpola.
\newblock {Mean teachers are better role models: Weight-averaged consistency
  targets improve semi-supervised deep learning results}.
\newblock {\em Advances in Neural Information Processing Systems},
  2017-December:1196--1205, 2017.

\bibitem{Toledo-cort}
Santiago Toledo-cort, Melissa De~La Pava, and Oscar Perd.
\newblock {Diabetic Retinopathy Diagnosis and Uncertainty}.

\bibitem{Voets2019}
Mike Voets, Mollersen Kajsa, and Lars Bongo.
\newblock {Reproduction study using public data of : Development and validation
  of a deep learning algorithm for detection of diabetic retinopathy in retinal
  fundus photographs}.
\newblock pages 1--11, 2019.

\bibitem{Vu2021}
Yen Nhi~Truong Vu, Richard Wang, Niranjan Balachandar, Can Liu, Andrew~Y. Ng,
  and Pranav Rajpurkar.
\newblock {MedAug: Contrastive learning leveraging patient metadata improves
  representations for chest X-ray interpretation}.
\newblock pages 1--14, 2021.

\bibitem{Xie2020}
Qizhe Xie, Zihang Dai, Eduard Hovy, Minh-thang Luong, and Quoc~V Le.
\newblock {Unsupervised Data Augmentation for Consistency Training}.
\newblock (NeurIPS):1--20, 2020.

\bibitem{Xie2017}
Saining Xie, Ross Girshick, Piotr Doll{\'{a}}r, Zhuowen Tu, and Kaiming He.
\newblock {Aggregated residual transformations for deep neural networks}.
\newblock {\em Proceedings - 30th IEEE Conference on Computer Vision and
  Pattern Recognition, CVPR 2017}, 2017-Janua:5987--5995, 2017.

\bibitem{Xie}
Yingpeng Xie, Qiwei Wan, Guozhen Chen, Yanwu Xu, and Baiying~Lei B.
\newblock {\em {Retinopathy Diagnosis Using Generative Adversarial Network}},
  volume~1.
\newblock Springer International Publishing.

\end{thebibliography}

\end{document}